\documentclass[runningheads]{llncs}
\usepackage{cite}
\usepackage{graphicx}
\usepackage{epstopdf}
\graphicspath{{eps/}}
\usepackage{upgreek}
\DeclareGraphicsExtensions{.eps}
\usepackage[cmex10]{amsmath}
\usepackage{amsfonts}
\usepackage{amssymb}
\usepackage{enumerate}
\usepackage{url}
\newcommand{\keywords}[1]{\par\addvspace\baselineskip
\noindent\keywordname\enspace\ignorespaces#1}
\usepackage{multirow}
\usepackage{siunitx}
\usepackage{scrextend}
\usepackage{tabularx}
\usepackage{fixltx2e}
\usepackage{subfigure}
\usepackage{floatrow}
\usepackage{wrapfig}
\usepackage{booktabs}

\begin{document}
\title{\emph{Gaze2Segment}: A Pilot Study for Integrating Eye-Tracking Technology into Medical Image Segmentation}
\titlerunning{Gaze2Segment}

%
%
\authorrunning{Khosravan et al.}

\author{Naji Khosravan$^1$, Haydar Celik$^2$, Baris Turkbey$^2$, Ruida Cheng$^2$,
Evan McCreedy$^2$, Matthew McAuliffe$^2$,  Sandra Bednarova$^2$, Elizabeth Jones$^2$, Xinjian Chen$^3$, Peter L. Choyke$^2$, Bradford J. Wood$^2$, Ulas Bagci$^{1,*}$}
%
%
\tocauthor{Naji Khosravan$^1$, Haydar Celik$^2$, Ismail Baris Turkbey$^2$, Ruida Cheng$^2$,
Evan McCreedy$^2$, Matthew McAuliffe$^2$,  Sandra Bednarova$^2$, Elizabeth Jones$^2$, Xinjian Chen$^3$, Peter Choyke$^2$, Bradford Wood$^2$, Ulas Bagci$^{1,*}$}
\institute{$^1$Center for Research in Computer Vision (CRCV), University of Central Florida (UCF), Orlando, FL.\\$^2$National Institutes of Health (NIH), Bethesda, MD.\\$^3$Soochow University, Suzhou City, China.\\$^*$:ulasbagci@gmail.com}
\maketitle

\begin{abstract}
This study introduced a novel system, called \emph{Gaze2Segment}, integrating biological and computer vision techniques to support radiologists' reading experience with an automatic image segmentation task. During diagnostic assessment of lung CT scans, the radiologists' gaze information were used to create a visual attention map. This map was then combined with a computer-derived saliency map, extracted from the gray-scale CT images. The visual attention map was used as an input for indicating roughly the location of a object of interest. With computer-derived saliency information, on the other hand, we aimed at finding foreground and background cues for the object of interest. At the final step, these cues were used to initiate a seed-based delineation process. Segmentation accuracy of the proposed \emph{Gaze2Segment} was found to be 86\% with dice similarity coefficient and $1.45$ mm with Hausdorff distance. To the best of our knowledge, \emph{Gaze2Segment} is the first true integration of eye-tracking technology into a medical image segmentation task without the need for any further user-interaction. 
\end{abstract}

\section{Introduction}
Eye-tracking based research in radiology can be categorized into two main groups. First group of research focuses on psychological viewpoint (e.g. examining attentional behavior) such as the early work of Just et al.~\cite{just1980theory}. Relevant to medical imaging field, most of these studies have been accomplished to understand radiologists' visual search patterns, differences of search patterns, and expert and non-expert visual search discriminations~\cite{mallett2014tracking}. The second group considers eye-tracking as an interaction tool with computers. For instance, Ware et al\cite{ware1987evaluation} used eye-tracker information as an input to perform a predefined task in the computer. In a different study, Sadeghi et al.~\cite{sadeghi2009hands} showed the advantage of using eye-tracking over using mouse clicks as an interaction tool for segmentation task in general. Despite significant advances in human-computer interaction, the use of eye-tracking technology to perform image analysis tasks in radiology remains \textit{largely} untouched. 

\begin{figure}
	\centering
		\includegraphics[scale=0.32]{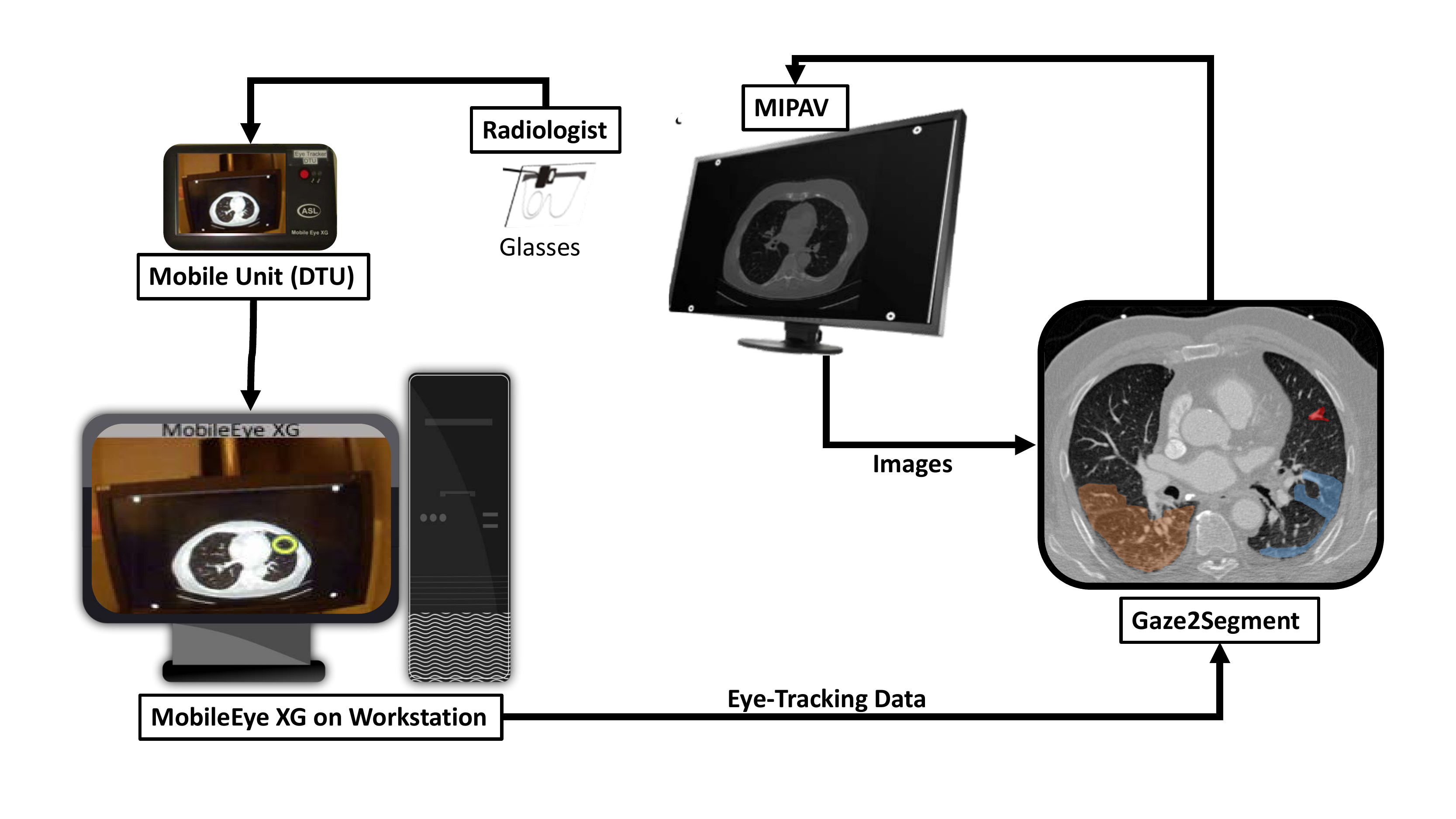}
	\caption{Overview of the integrated eye-tracking and \textit{\textbf{Gaze2Segment}} system}
	\label{fig:eyetrackingsystem}
\end{figure}
\begin{figure}[h]
		\includegraphics[scale=0.27]{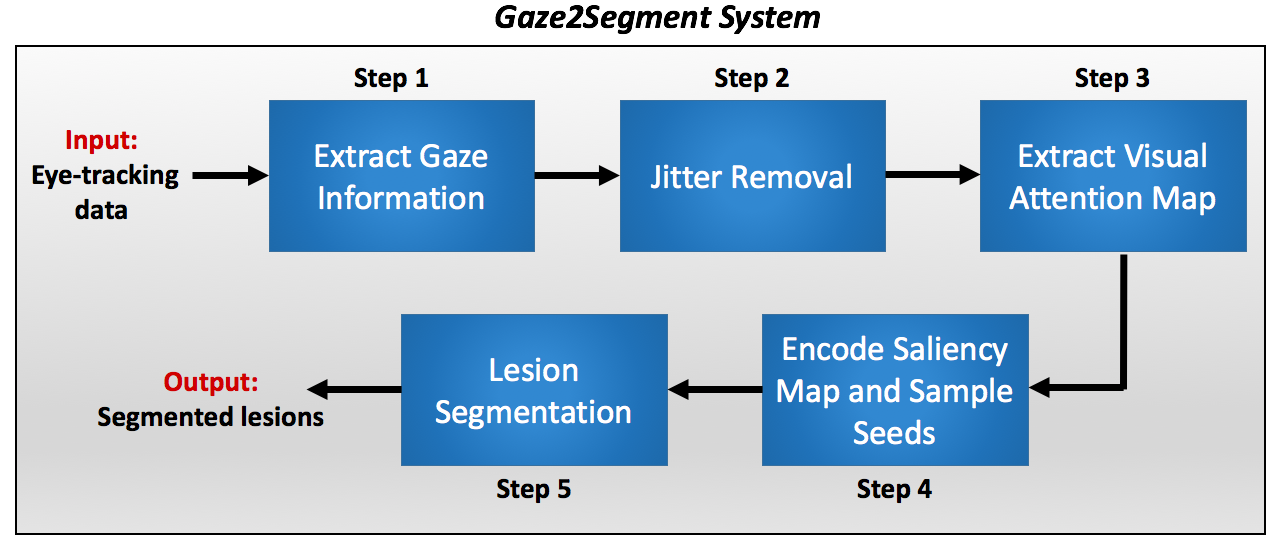}
	\caption{\textit{\textbf{Gaze2Segment}} has five steps to perform a segmentation task. Input is inferred from the eye-tracking data (see Figure 1).\label{fig:overview}}
\end{figure}

In this study, we propose a pilot system that uses gaze information from the eye-tracker as an input to perform \textit{a fully automatic} image segmentation for radiology scans. To the best of our knowledge, this is the first study integrating biological and computer vision methods synergistically to conduct a quantitative medical image analysis task. The integration of the proposed algorithm, \textit{\textbf{Gaze2Segment}}, into the eye-tracker system is illustrated in Figure~\ref{fig:eyetrackingsystem}.
Our motivation for the use of eye-tracking in medical image segmentation task lies in the following facts. The segmentation process includes two relevant (and complementary) tasks: \textit{recognition} and \textit{delineation}~\cite{bagci2012hierarchical}. While delineation is the act of defining the spatial extent of the object boundary in the image, recognition (i.e., localization or detection) is the necessary step for determining roughly where the object is. Automatic recognition is a difficult task; hence, manual or semi-automated methods are often devised for this purpose. Available automatic recognition methods usually employ an exhaustive search or optimization. \textit{We postulate herein that eye-tracking can be used as an effective recognition strategy for the medical image segmentation problems.} Towards this aim, we developed the \textit{\textbf{Gaze2Segment}} consisting of the following five major steps, as illustrated in Figure~\ref{fig:overview}:
\begin{itemize}
  \item \textbf{Step 1:} Real-time tracking of radiologists' eye movements for extracting gaze information and mapping them into the CT scans (i.e., converting eye tracker data into image coordinate system).
  \item \textbf{Step 2:} Jitter Removal for filtering out the unwanted eye movements and stabilization of the gaze information.
  \item \textbf{Step 3:} Creating visual attention maps from gaze information and locating object of interest from the most important attention regions.
  \item \textbf{Step 4:} Obtaining computer-derived local saliency and gradient information from gray-scale CT images to identify foreground and background cues for an object of interest.
  \item  \textbf{Step 5:} Segmenting the object of interest (identified in step 3) based on the inferred cues (identified in Step 4).
\end{itemize}

\section{Method}
\subsection*{Step 1: Eye-Tracking and Extracting Gaze Information}
We have used MobileEye XG eye-tracker technology (ASL, Boston, MA) to build our system (Figure~\ref{fig:eyetrackingsystem}). This device has  eye and scene cameras that attach to the glasses (or an empty frame in case the participating radiologist already has eye glasses). The two cameras are adjustable to fit different users' settings. While the eye camera records the eye movements, the scene camera (second camera, directed forward) records the monitor being observed by the radiologist at $60$Hz of data rate. The eye camera monitors the pupil orientations and reflective angle using corneal reflection of $3$-infrared dots on the eye from a reflective mirror. These dots are transparent to visible spectrum and nothing obscures the radiologists' field of view. The data from these two cameras were transferred into a workstation through a mobile display/transmit unit using an ethernet cable in real-time. Then, points of gaze were computed on the scene video, which was recorded at 60 frames per second. A calibration needs to be performed by the radiologist before every image reading experiment to match the eye movement data and the $640\times 480$ scene video. The system outputs pupil diameter and gaze coordinates with respect to the scene camera's Field Of View (FOV) on a \textit{.csv} file with timestamps (Figure~\ref{fig:eyetrackingsystem}). Once the calibrated gaze coordinates, scene video, and timestamp were created, gaze coordinates on the scene video ($g^{v}$) were converted onto the gaze coordinates on the stimulus ($g^{s}$). 

Our pilot study focuses on a realistic scan evaluation by a radiologist without inserting any environmental or psychological constraints. As a part of this realistic experiment in a dark reading room, we have collected chest CTs pertaining to patients diagnosed with lung cancer. Unlike relatively simpler experiments with X-Rays, there are numerous slices to evaluate in 3D CTs. In addition, radiologists may visit the same slice more than once during their reading, including changing the image views into axial, coronal, and sagittal sections. To mitigate these, an image viewer plugin was developed to be integrated into the open source MIPAV image analysis software \cite{mcauliffe2001medical}. The plugin simply records mouse manipulations including scrolling, contrast change, and button clicks with the associated timestamps.

\subsection*{Step 2: Jitter Removal and Gaze Stabilization}
Eye-tracking data naturally contains jittery noises. While looking at a single object, users normally believe that they look at the object steadily. However, eyes have small jittery movements that causes the gaze location to be unstable. Using such noisy data can create uncertainties in image analysis tasks. In order to remove jitter, while preserving global gaze patterns, a new smoothing operator ($J$) was formulated as follows. Since gaze coordinates on the stimulus ($g^{s}$) included a set of points on $xy$-coordinate system (i.e., planar), Euclidean distance between any consecutive coordinate points could be used for smoothing as values that fall within the small distance neighborhood were eliminated:  $$\text{if } || g^{s}(i)-g^{s}(i+1)|| \leq \varepsilon,$$ then, $g^{s}(i)$ is set to $g^{s}(i+1)$, where $i$ indicates the gaze points in an order they have been looked at by the user, and $\varepsilon$ was a pre-defined distance (based on the empirical evaluation of experimental data) and set as 7.5 mm, meaning that all the pixels within $\varepsilon$-neighborhood of $i$ were considered to be pertaining to the same attention regions. 


\subsection*{Step 3: Visual Attention Maps}
There are two major visual search patterns identified so far that radiologists normally follow for reading volumetric radiology scans: \textsl{drilling} and \textsl{scanning}\cite{drew2013scanners}. While \textsl{drillers} spend less time on a single area in an image slice and tend to scroll fast between slices (backward and forward), \textsl{scanners} spend more time on examining a single slice and then move to the next slice. Thus, it's a valid hypothesis that radiologists spend more time on the regions that are more suspicious to them. Hence, the possibility of abnormality presence in those areas is higher compared to the other regions. This fact can be used to perform an image analysis task in suspicious areas of radiology scans. %

Considering the above mentioned information, as well as the theory of \textit{Von Helmholtz}, claiming that eye movements reflect the will to inspect interesting objects in fine detail although visual attention can still be consciously directed to peripheral objects\cite{von2005treatise},  we used the time information (from timestamp on the data) to create visual attention map by encoding the regions to which radiologists divert their attention more than other regions. The time spent on a specific area might be different between drillers and scanners but the time that is spent on potentially abnormal areas is still \textit{relatively higher} than other areas for a specific user regardless of the search method. These reasons make the time a reliable factor to derive an attention map. 

For each gaze point on the stimulus $g^{s}(i)$, an attention value $a(i)\in [0,1]$ was created by mapping the corresponding timestamp $t(i)$  of the gaze coordinate in piece-wise linear form as follows:
\begin{equation}
a(i)=\left\{\begin{array}{lr}
\frac{t(i) - \hat{t} }{t_{max}-\hat{t}}, & t(i) > \hat{t},\\
0, & \textrm{otherwise,}
\end{array} \right.
\label{eqn:threshold}
\end{equation}
where $t_{max}={argmax_i }$ $t(i)$ and $\hat{t}$ can be set into 0 in order to assign an attention value for every gaze coordinate. For practical reasons, since many gaze coordinates may have very small timestamps (i.e., in milliseconds), those gaze coordinates can be removed from the analysis by setting a larger $\hat{t}$.

\subsection*{Step 4: Local Saliency Computation for sampling Foreground/Background Cues}
In biological vision, humans tend to capture/focus on most salient regions of an image.  In computer vision, many algorithms have been developed to \textit{imitate} this biological process by defining a \textit{saliency} concept with different context. The mostly used definition of saliency is based on the distinctiveness of regions with respect to their local and global surroundings. Although this definition is plausible for many computer vision tasks, it alone may not be suitable for defining salient regions in radiology scans where object of interests are not often as distinctive as expected. In addition, radiologists use high level knowledge or contextual information to define regions of interest. Due to all these reasons, we propose to use a \textit{context-aware saliency} definition that aims at detecting the image regions based on contextual features\cite{goferman2012context}. In our implementation, we extracted image context information by predicting which point attracts the most attention. This step combines radiologist's knowledge with image context. The context-aware saliency explains the visual attention with feature-driven four principles, three of which were implemented in our study: (1) local low-level considerations, (2) global considerations, (3) visual organization rules, and (4) high-level factors. 

(1) For local low-level information, image was divided into local patches ($p_u$) centered at pixel $u$, and for each pair of patches, their distance ($d_{position}$) and normalized intensity difference ($d_{intensity}$) were used to assess saliency of a pixel $u$, as formulated below:

\begin{equation}
d(p_{u},p_{v})=d_{intensity}/(1+\lambda d_{position}),
\label{eqn:saliency}
\end{equation}
where $\lambda$ is a weight parameter. Pixel $u$ was considered \textit{salient} when it was highly dissimilar to all other image patches, $d(p_{u},p_{v})$ is high $\forall v$. 

(2) For global considerations, a scale-space approach was utilized to suppress frequently occurring features such as background and maintain features that deviate from the norm. Saliency of any pixel in this configuration was defined as the average of its saliency in $M$ scales $\lbrace (r_{1},r_{2},...,r_{M}), r\in R \rbrace$: 
\begin{eqnarray}
\bar{S_{u}}=(1/M)\sum _{r\in R}S_{u}^{r}\\
S_{u}^{r}=1-exp\lbrace-(1/K)\sum _{k=1}^{K}d(p_{u}^{r},p_{v}^{r})\rbrace & \text{ for} & (r\in R).
\end{eqnarray}
This scale-based global definition combined $K$ most similar patches for the saliency definition and indicated a more salient pixel $u$ when $S_{u}^{r}$ was large.

(3) For visual organization rules, saliency was defined based on the Gestalt laws suggesting areas that were close to the foci of attention should be explored significantly more than far-away regions. Hence, assuming $d_{foci}(u)$ is the Euclidean distance between pixel $u$ and the closest focus of attention pixel, then the saliency of the pixel was defined as $\hat{S_{u}}=\bar{S_{u}}(1-d_{foci}(u))$. A point was considered as a focus of attention if it was salient.

(4) High-level factors such as recognized objects can be applied as a post processing step to refine saliency definition. In our current implementation, we did not apply this consideration.

Since we inferred \emph{where} information of object of interest from visual attention map (Step 3), we only explored \emph{what} part of object of interest from saliency definition. Once saliency map was created, we confined our analysis into the regions indicated by corresponding visual attention maps ($a(u)$). Since saliency map included object of interest information, we extracted foreground information from this map (called foreground cues/seeds) by simply setting the most salient pixel in this region as a foreground cue. This step helped relocating the attention gaze exactly to the center of the closest most salient object and allowed a perfect seed selection. 

\begin{figure}
	\centering
		\includegraphics[scale=0.35]{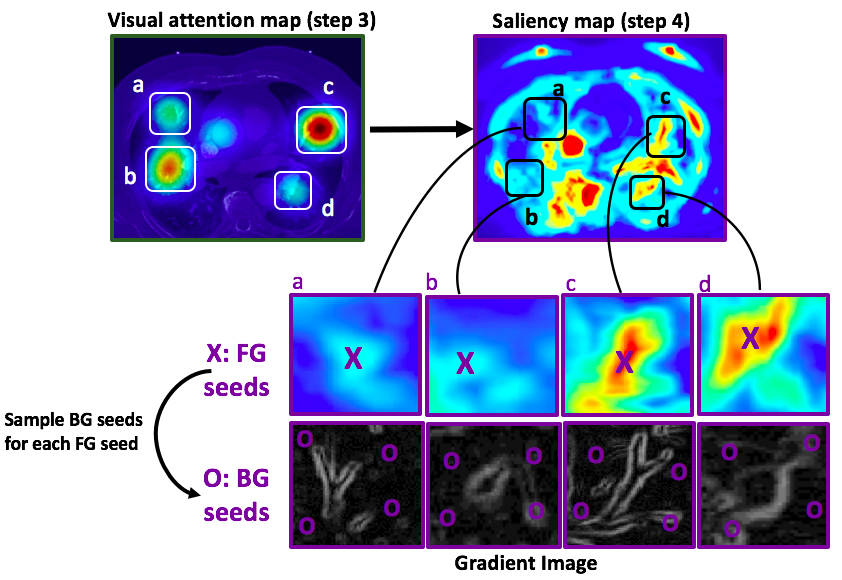}
	\caption{Foreground (FG) regions were obtained from visual attention maps processed from gaze information. After this \textit{recognition} step, we identify most distinct FG seed by using the corresponding regions of saliency map. Once FG seeds are allocated, background (BG) seeds are found by using gradient information of the gray-scale CT image. For each FG seed, four perpendicular directions are searched and edge locations are used to select BG seeds. \label{fig:seed}}
\end{figure}

Furthermore, we defined the background cues for a given local region, indicated by the visual saliency map, as follows. We first computed the gradient information $\nabla I$ from a gray-scale CT image $I$. For a given visual attention map $a(u)$ and saliency map $S(u)$ at the pixel $u$,  we employed a search starting from $\nabla I(u)$ and moving into 4 perpendicular directions. Our search was stopped soon after we passed through a high intensity value on the gradient image because object boundaries show high gradient values. Those four pixels found outside the object boundary were considered as background cues. This process is illustrated in Figure~\ref{fig:seed}. 

\subsection*{Step 5: Lesion Segmentation}
After identifying background and foreground seeds, any seed-based segmentation algorithm such as graph-cut, random walk (RW), and fuzzy connectivity,  can be used to determine precise spatial extent of the object of interest (i.e., lesion). In our work, we choose to implement RW as it is fast and robust, and offers optimal image segmentation for a given set of seed points. Details of the conventional RW image segmentation algorithm can be found in~\cite{grady2006random}.

\section{Results}
We tested our system on four chest CT volumes pertaining to patients diagnosed with lung cancer, evaluated by three radiologists having different levels of expertise. In-plane resolution of the images is $512 \times 512$ with a voxel size of $0.58\times0.58\times1.5$ $mm^3$. Imaging data and corresponding lesion labels as well as annotations were obtained from Lung Tissue Research Consortium (LTRC) (\url{https://ltrcpublic.com/}) with an institutional agreement. Blind to diagnostic information of the chest CT scans, the radiologists read the scan once, and interpret the results in routine radiology rooms. Participating radiologists have more than 20, 10, and 3 years of experiences, respectively. This variability in experience levels allowed us to test robustness of our system. As shown by results regardless of user experience and pattern of gaze and attention, our system perfectly captured the attention gaze locations and performed the segmentation successfully.

\begin{figure}
	\centering
		\includegraphics[scale=0.35]{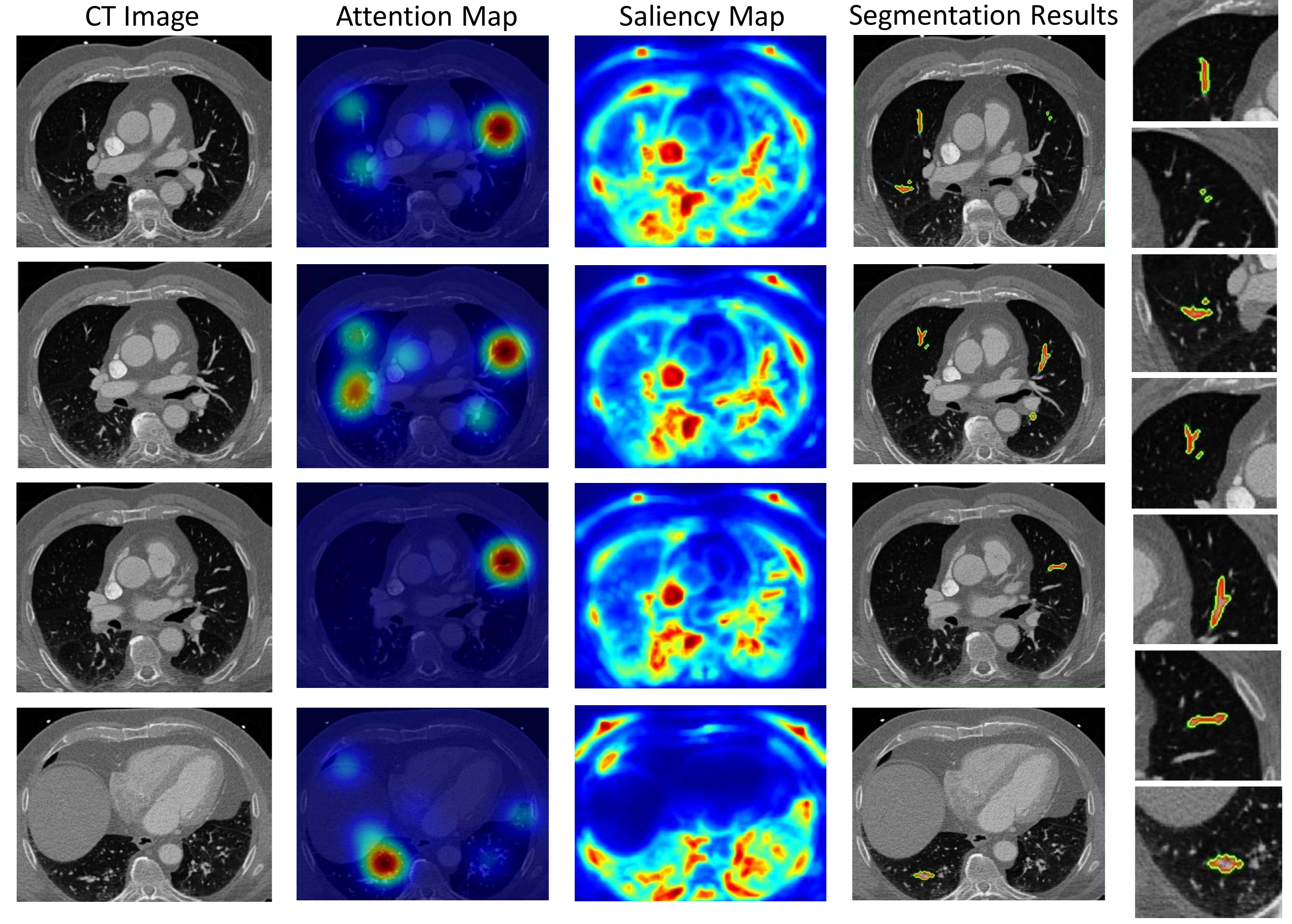}
	\caption{Qualitative evaluation of medical image segmentation through  \textit{\textbf{Gaze2Segment}} system is illustrated. Last column shows the segmentation results zoomed in for better illustration. \label{fig:qualitative}}
\end{figure}

Figure~\ref{fig:qualitative} shows the proposed system's visual attention map, local saliency map,  foreground/background seed samples, and segmentation results at different anatomical locations. For quantitative evaluation of segmentation results, we used reference standards from LTRC data set in addition to  an independent evaluation by one of the participating radiologists (through manual annotations). We have used dice similarity coefficient (DSC) and Haussdorff Distance (HD) to evaluate accuracy of segmentation results on two reference standards. The average DSC was found to be $86\%$ while average HD was $1.45$ mm. We did not find statistically significant difference between segmentation results when manual seeding and interactive RW were used (t-test, $p>0.05$).

\begin{figure}
	\centering
		\includegraphics[scale=0.650]{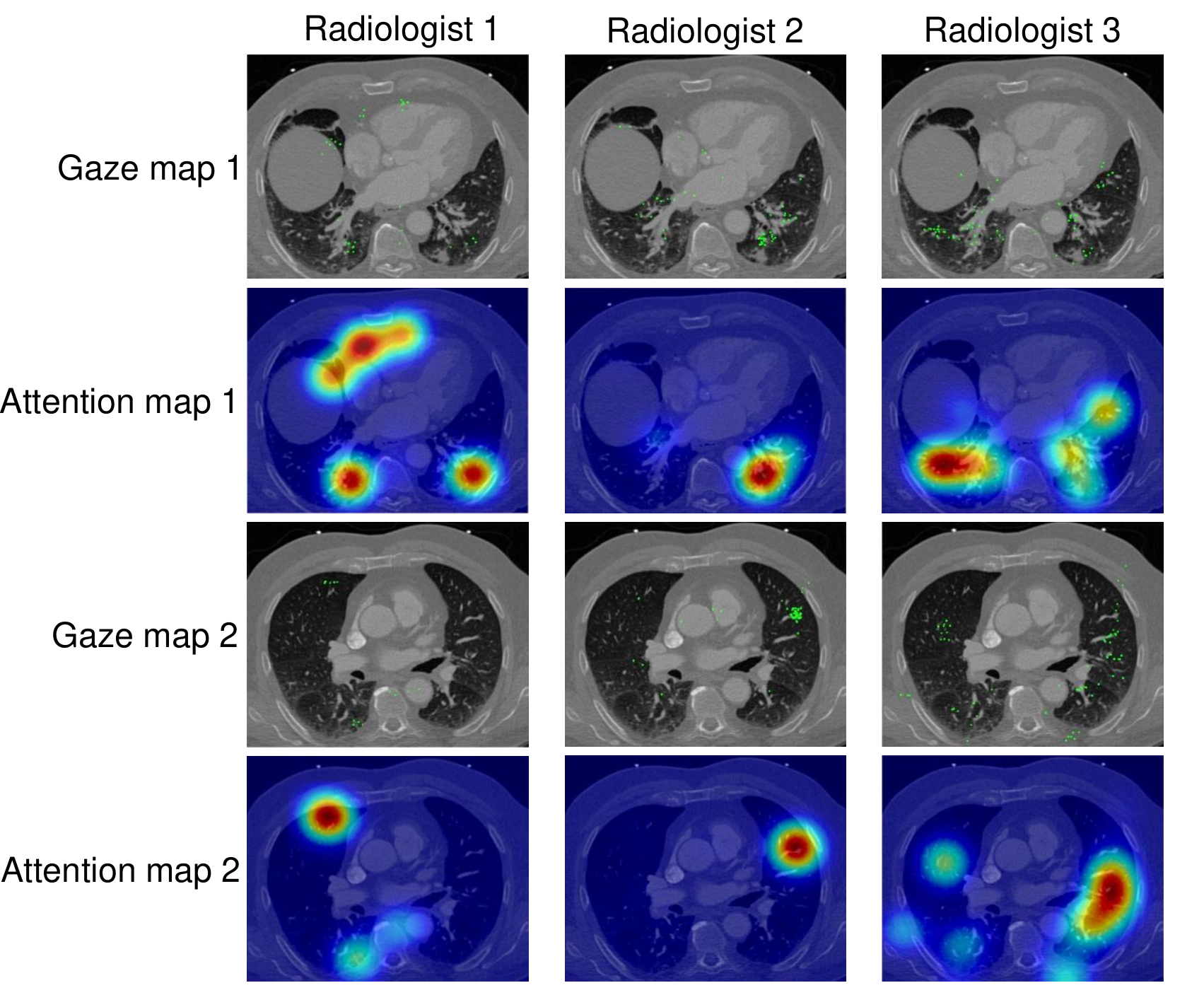}
	\caption{Qualitative comparison of attention and gaze maps is illustrated. \label{fig:gaze}}
\end{figure}

\begin{figure}
	\centering
		\includegraphics[scale=0.65]{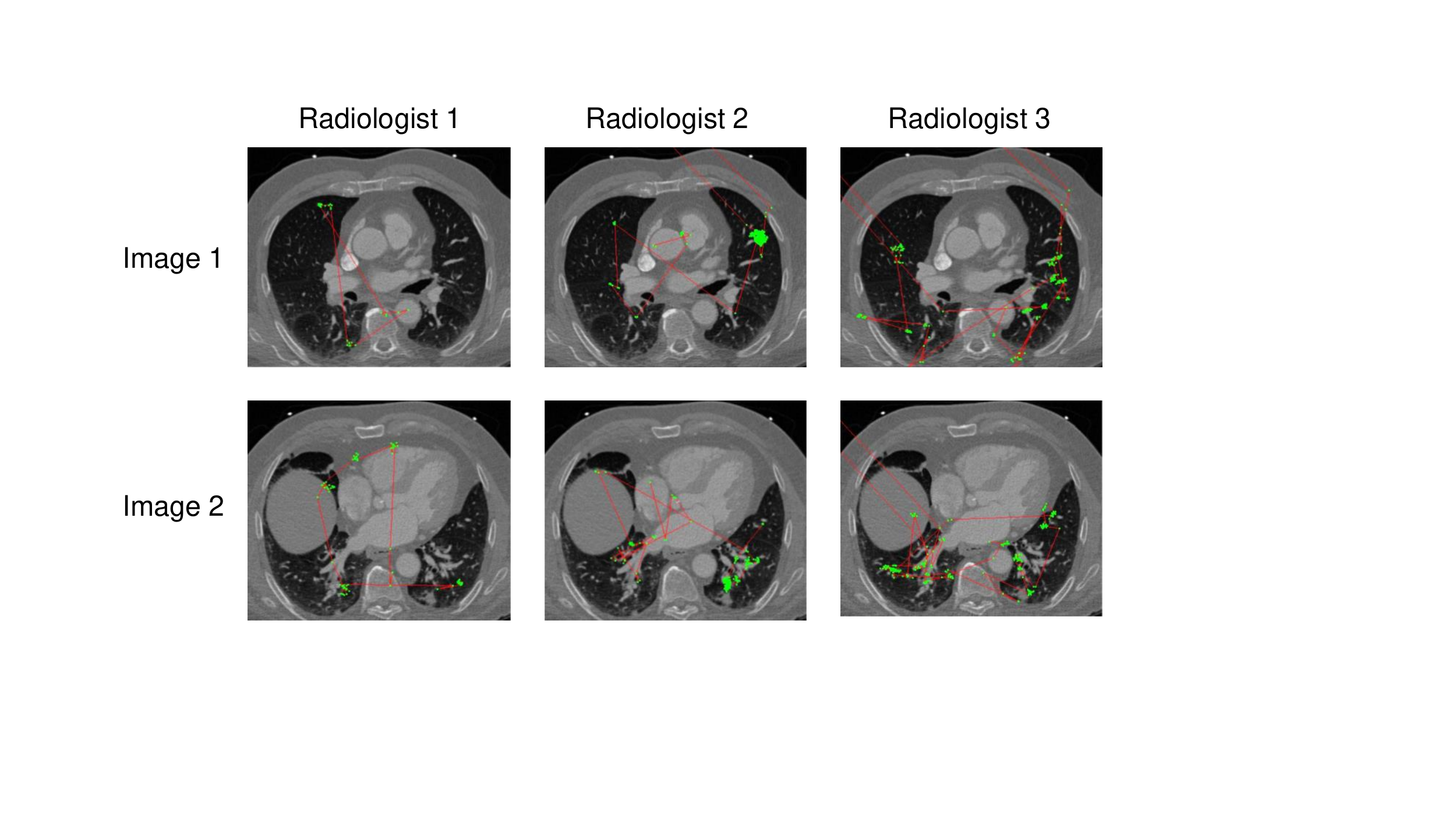}
	\caption{Comparisons of gaze patterns pertaining to participating radiologists. Second image has district pathological regions, having overlapped attentional points by all radiologists while first image has distinct attentional patterns. \label{fig:pattern}}
\end{figure}

Figure~\ref{fig:gaze} shows a comparison of gaze and attention maps for two sample slices of the chest CT volume screened by participating radiologists. How the attentional points are distributed over the lung CT volume is arguable based on the experience levels of the radiologists. As Figure~\ref{fig:gaze} illustrates, the less experienced the radiologist is(radiologist 3), the larger the volume of search is compared to the expert radiologists (radiologist 1 and 2).  For the selected slices, radiologists' gaze patterns are mapped on the images to compare radiologists' attention patterns in Figure~\ref{fig:pattern}. While attention/search patterns seem to be distinct in the first image, pathological regions (in the second image) have overlapped attentional points among radiologists.

\section{Discussion}
Since our work is a pilot study, there are several limitations that should be noted. First, we used a limited number of imaging data to test our system. However, it should be noted that gathering a large number of imaging data with corresponding eye-tracking information is a time consuming task. With that said, our team is working on gathering more imaging and eye-tracking data to extend experiments for our future studies. Second, there were several region of interests (non-lesion based) identified and segmented with the proposed system as potential lesion locations. It is because the visual attention information indicated that the radiologists spent several seconds on those regions, and our system naturally considered those regions as potential lesion locations. This can be solved by two ways: allowing the expert to eliminate those false positives interactively, or including a computer-aided detection system helping to remove such objects automatically.

Third, although the proposed system is derived from the solid theory of biological and computer vision, there may be additional computational tunings necessary. When different organs and imaging modalities are in consideration for a similar radiology reading experience, methods presented herein should be trained and tuned based on the imaging characteristics and saliency definition. Despite the challenges that might appear due to modality differences, our system has the potential for addressing such difficulties. Fourth, the system parameters such as $\varepsilon$ or $\hat{t}$ are selected empirically. A more reliable and data-driven approach could replace this manual step. Fifth, the segmentation is performed off-line after the data is recorded. Performing the whole process online and during the reading experience in radiology rooms is the future goal. Our initial results on this pilot study add sufficient evidences towards this realistic and innovative goal.

\section{Conclusion}
In this paper, an automated eye-tracking system was integrated into a medical image segmentation process. For this task, we have successfully combined biological and computer vision techniques for the first time in radiology scan reading setting. We used radiologist's gaze information to extract visual attention map and then complement this information with the computer derived local saliency information from radiology images. By utilizing these two information, we first sampled object and background cues from a region of interest indicated by the eye-tracking and performed a medical image segmentation task. By this way, we proved that gaze information can be used effectively to address the recognition problem of image segmentation, causing a real-time quantification of radiology scans. Our main contribution is to combine biological vision and attention information with image context through saliency map. This has been achieved to perform a quantitative analysis of medical scans during the reading experience and without the need for any further interaction from the user side.

\bibliographystyle{unsrt}
\bibliography{gaze2segment}
\end{document}